\documentclass[letterpaper]{article} % DO NOT CHANGE THIS
\usepackage{aaai24}  % DO NOT CHANGE THIS
\usepackage{times}  % DO NOT CHANGE THIS
\usepackage{helvet}  % DO NOT CHANGE THIS
\usepackage{courier}  % DO NOT CHANGE THIS
\usepackage[hyphens]{url}  % DO NOT CHANGE THIS
\usepackage{graphicx} % DO NOT CHANGE THIS
\usepackage{natbib}  % DO NOT CHANGE THIS AND DO NOT ADD ANY OPTIONS TO IT
\usepackage{caption} % DO NOT CHANGE THIS AND DO NOT ADD ANY OPTIONS TO IT
\usepackage{algorithm}

\usepackage{algorithmicx}
\usepackage{algpseudocode}

\usepackage{tikz} % nice language for creating drawings and diagrams
\usepackage{todonotes}
\usepackage{amsmath,amsfonts,mathtools,amsthm,amssymb}
\usepackage{IEEEtrantools}

\usepackage{newfloat}
\usepackage{listings}
\DeclareCaptionStyle{ruled}{labelfont=normalfont,labelsep=colon,strut=off} % DO NOT CHANGE THIS
\floatstyle{ruled}
\newfloat{listing}{tb}{lst}{}
\floatname{listing}{Listing}

\definecolor{myBlue}{rgb}{0,0.2,1.0}
\definecolor{myBlack}{rgb}{0.0,0.0,0.0}
\newcommand{\chao}[1]{{\color{myBlack}#1}}

\newcommand{\Lim}[1]{\raisebox{0.5ex}{\scalebox{0.8}{$\displaystyle \lim_{#1}\;$}}}

\newtheorem{theorem}{Theorem}
\newtheorem{lemma}{Lemma}

\title{Monte Carlo Tree Search in the Presence of Transition Uncertainty}
\author{
    Farnaz Kohankhaki\textsuperscript{\rm 1}\equalcontrib,
    Kiarash Aghakasiri\textsuperscript{\rm 1, \rm 2}\equalcontrib,
    Hongming Zhang\textsuperscript{\rm 1},
    Ting-Han Wei\textsuperscript{\rm 1},
    Chao Gao\textsuperscript{\rm 2},
    Martin Müller\textsuperscript{\rm 1}
}
\affiliations{
    \textsuperscript{\rm 1}University of Alberta,
    \textsuperscript{\rm 2}Edmonton Research Center, Huawei Canada\\
    \{kohankha, aghakasi, hongmin2, tinghan, mmueller\}@ualberta.ca,
    cgao3@outlook.com
}

\begin{document}

\maketitle

\begin{abstract}
Monte Carlo Tree Search (MCTS) is an immensely popular search-based framework used for decision making. It is traditionally applied to domains where a perfect simulation model of the environment is available. We study and improve MCTS in the context where the environment model is given but imperfect. We show that the discrepancy between the model and the actual environment can lead to significant performance degradation with standard MCTS. We therefore develop Uncertainty Adapted MCTS (UA-MCTS), a more robust algorithm within the MCTS framework. We estimate the transition uncertainty in the given model, and direct the search towards more certain transitions in the state space. We modify all four MCTS phases to improve the search behavior by considering these estimates. We prove, in the corrupted bandit case, that adding uncertainty information to adapt UCB leads to tighter regret bound than standard UCB. Empirically, we evaluate UA-MCTS and its individual components on the deterministic domains from the MinAtar test suite. Our results demonstrate that UA-MCTS strongly improves MCTS in the presence of model transition errors.\footnote{\url{https://github.com/ualberta-mueller-group/UAMCTS}}

\end{abstract}

\section{Introduction}
The Monte Carlo Tree Search (MCTS) framework \citep{MCTSSurvey} approaches sequential decision-making problems by selective lookahead search. It manages the balance of exploration and exploitation with techniques such as UCT \citep{uct}.
Often combined with machine learning, it has been enormously successful in both games \citep{AlphaGo,MIMHex,MoHex,YOPT,checkers} and non-game applications \citep{FA,HOOT,MIP,MEG}. 
In these applications, a perfect simulation model allows for efficient lookahead search.
However, in many practical applications, only an imperfect model is available to the agent. Yet lookahead using such a model can still be useful. We improve MCTS for this setting.

One research area that studies imperfect models of the environment is model-based reinforcement learning (MBRL). Here, an agent builds its own model through limited real world interactions.
The resulting learned model, when used for lookahead search, can either be for planning or for producing more accurate training targets \citep{silver2008sample}. It can also be used to generate simulated training samples for better sample efficiency \citep{richbook}.
The learned model may be inaccurate for many reasons, including stochasticity of the environment, insufficient training, insufficient capacity, non stationary environments, etc. 
%Model-based methods can also be used for decision-time planning, which is useful in cases where the value function estimators have low capacity so the temporary values coming from the search methods can compensate for that \cite{silver2008sample}. 
Consequently, there is a rich body of research on \emph{uncertainty} in MBRL \citep{zaheer,AdaMVE,steve}.

While previous approaches to using search with imperfect models exist \citep{cmax,cmax++}, to the best of our knowledge, there is no prior work that directly adapts MCTS to deal with model uncertainty.
In our work, we define transition uncertainty as a measure of difference between the state transitions in the perfect model and in the model that is available to the agent. 
We use a neural network to estimate this uncertainty.

Our Uncertainty Adapted MCTS (UA-MCTS) approach implements the main components of the MCTS framework in a way that guides the search away from states with high uncertainty. We compare the performance of our proposed methods with MCTS baselines in three deterministic MinAtar environments \citep{minAtar}.
In each case the search agent ``believes'' it is playing the real game. However, the rules of the game itself have changed, and the agent only learns about this change slowly when it acts in the real environment.
The results show that UA-MCTS is able to outperform the baseline MCTS with an imperfect model. 

Our approach is inspired by the work of \citep{cmax} where a robotic arm has to solve tasks despite being handicapped, e.g. by a broken motor or by an unmodeled weight restriction. To show how an agent should adapt UCB-based exploration strategy in the presence of environment uncertainties, we first consider a case of stochastic bandits~\cite{lattimore2020bandit} along with corrupted feedback. We prove that incorporating uncertainty information can enhance the performance of UCB, yielding a regret bound that is more constrained compared to the standard UCB. We also prove that in the general case of tree search, with similar modification of UCT, our UA-MCTS approach maintains its completeness property, ensuring that as the number of iterations goes to infinity, all nodes will be consistently explored.
To further motivate our approach, we compare the scenarios of learning to improve the transition function, using MCTS, directly against the easier task of just learning a transition uncertainty function with UA-MCTS. In both cases, learning occurs online; the former is used with MCTS while the latter is used with UA-MCTS. Our results show that learning the transition function is much harder than learning transition uncertainty, which justifies the use of UA-MCTS in such settings.

\section{Background}
\subsection{The Imperfect Model Problem Setting}
We focus on \textit{deterministic} Markov Decision Processes (MDP) which can be expressed as a 5-tuple $(\mathcal{S},\mathcal{A},\mathcal{R},M,\gamma)$ with
state space $\mathcal{S}$, action space $\mathcal{A}$, rewards $\mathcal{R}$ which
map an $(s,a)$ pair to a scalar reward,
deterministic dynamics $s_{t+1} = M(s_t,a)$, and a
discount factor $0\leq\gamma\leq1$.
An agent \textit{policy} maps each state in $\mathcal{S}$ to an action in $\mathcal{A}$. The goal is to find a policy that maximizes the total discounted reward.

During the course of a search, the agent only has access to an \textit{imperfect model} $\hat{M}(s,a)$ of the environment.
We assume that $\hat{M}$ is both given and fixed within one search. This is in contrast to the model learning case where $\hat{M}$ is neither given nor fixed. Each search is used to select an action, which is then executed in the ``real world'' $M$.

Both $M$ and $\hat{M}$ take a state and an action as input and return the next state. However, the next state predicted by $\hat{M}$ might be incorrect. 
After each search in $\hat{M}$, the agent decides on an action $a$, which is then executed in the ``real world'' $M$, and observes the next true state. $\hat{M}$ and $M$ share the same $\mathcal{S}$, $\mathcal{A}$, $\mathcal{R}$, and $\gamma$.

We model the \textit{transition uncertainty} $U(s,a)$ as an (arbitrary) function of the difference between the next-state predictions made by $\hat{M}$ and $M$ for the same state-action pair $(s,a)$. Like $M$, $U$ is not directly available to the agent, thus it learns an estimation of $U(s,a)$, noted as $\hat{U}(s,a)$, during interactions with the environment. \chao{ A special case is that when the MDP contains only one state, the problem becomes a \emph{corrupted $K$-armed stochastic bandit} where there are $K$ actions. In this case, the \emph{transition uncertainty} leads to corrupted reward distribution ---  the agent takes arm $i$, instead of receiving from the reward from real distribution $p_i$, a reward is sampled from corrupted distribution $\hat{p}_i$. Following standard bandit assumptions, we use $\mu_i \in [0,1]$ and $\hat{\mu}_i \in [0,1]$ to denote the expected rewards for $p_i$ and $\hat{p}_i$, respectively. The difference is noted as $\delta_i = |\mu_i - \tilde{\mu}_i|$.     
}
% However, after each search, an action $a$ is executed from some state $s$ in the real world. From these samples of $U$, based on the difference between the real next state and the next state predicted by $\hat{M}$, the agent learns an estimate $\hat{U}(s,a)$ \textbf{many question marks on this sentence?}. 
% We acknowledge that the term uncertainty in the literature has been used for online model learning; however, in this work, the model $\hat{M}$ is considered fixed. A more accurate term could be model error but for the rest of this work, we use ``uncertainty'' for simplification. 

\subsection{Monte Carlo Tree Search (MCTS)}
\label{subsec:mcts}
MCTS conducts a selective tree search by repeating four steps (Algorithm \ref{alg:mcts}):
An in-tree \textit{selection} method such as UCT \citep{uct} descends along a path in the tree until a leaf node is reached,
which is then \textit{expanded}. A \textit{simulation} evaluates the selected leaf node. Finally, the nodes along the selection path, up to and including the root of the tree, are updated according to the evaluation result during \textit{backpropagation} \citep{MCTSSurvey}.
Each of the four steps of MCTS are modified in UA-MCTS.

We use the following notation: For node $v$ in the current search tree $T$, $S(v)$ is the feature vector of the state that is represented by $v$; $N(v)$ is the number of times $v$ has been visited throughout the search; $Q(v)$ is the sum of rewards observed at $v$;
$Par(v)$ is the parent of $v$ in $T$, with $Par(root)$ = \textit{NULL};
$Ch(v)$ is the current set of $v$'s children in $T$; and $\hat{U}(v)$ is the estimated transition uncertainty of $v$ which is equal to $\hat{U}(s,a)$, where $(s,a)$ is the transition in $T$ that leads to node $v$. The choices for $\hat{U}(v)$ are further elaborated on in Subsection \ref{subsec:learn-u}.
UA-MCTS is controlled by five hyperparameters: the total number of iterations $N_I$, the number $N_S$ of simulations to evaluate a leaf node of $T$, the maximum depth of each simulation $D_S$, the exploration constant $c$, and the uncertainty factor $\tau$ which controls how much the uncertainty affects the behavior of UA-MCTS.
\begin{algorithm}[tb]
\caption{MCTS Framework}\label{alg:mcts}
\begin{algorithmic}
\Function{MCTS}{$s_0$}
    \State create a root node $v_0$ with state $s_0$
    \For{$N_I$}
    \State $v_s \gets$ \Call{Select}{$v_0$}
    
    \If{$N(v_s) >  0$}
        \State $v_s \gets$ \Call{Expand}{$v_s$}
    \EndIf
    \State $value \gets \Call {Simulate}{S(v_s)}$
    \State \Call {Backpropagate}{$v_s$, $value$}
    \EndFor  
    \State $v_{best}\gets$ choose the most visited child of $v_0$
    \State \Return $action(v_{best})$
\EndFunction
\end{algorithmic}
\end{algorithm}
\section{Uncertainty Adapted UCB}
\chao{
Before diving into full MCTS for corrupted MDPs, we first consider the corrupted bandit case. For the real and corrupted environments, we assume the optimal arms are respectively noted as $i^*$ and $\hat{i}^*$. At time $t$ ($t>1$), let $\hat{x}_{i,(t-1)}$ be the empirical average reward collected for arm $i$ before time $t$, and $N_i(t-1)$ be the visit count of arm $i$. Naturally, the agent has to consider uncertainty information $\delta_i$; one sensible idea is to adapt the UCB score function for each arm $i$, as follows:   

\begin{equation*}
    \mathit{UA-UCB}_i(t) = \hat{x}_{i,t-1} + c \sqrt{\dfrac{\ln{(t-1)}}{N_{i,t-1}}} \cdot (1-\delta_i)
\end{equation*}
Here, $\hat{x}_{i,t}$ is an estimate for $\hat{\mu}_i$, e.g., empirical mean reward. %We can see when $\delta_i = 0$, UA-UCB becomes ordinary UCB, which is sensible since there is no difference between $\hat{p}_i$ and $p_i$.  
\subsection{Regret Bound}
We provide a regret bound for the UA-UCB strategy for corrupted bandits (proof shown in Appendix).
\begin{theorem}
   \begin{IEEEeqnarray*} {lCl}
    R_n &\leq&  \sum_{i=1}^{k} \Bigl[ \frac{4\beta_i}{\hat{\Delta}_i ^ 2} + C \Bigr] (\hat{\Delta}_i + \delta_i + \mu_{i^*} - \hat{\mu}_{\hat{i}^*})  \ln (n-1) \nonumber \\
\end{IEEEeqnarray*} \label{eq:regret bound new}
Here $C$ is a constant, $\beta_i = c^2 (1-\delta_i)^2$, $\Delta_i = \mu_{i^*} - \mu_{i}$ and $\hat{\Delta}_i = \hat{\mu}_{i^*} - \hat{\mu}_i$, assuming $0 \leq \delta_i \leq 1 - \sqrt{\frac{1}{2c^2}}.$
\end{theorem}
\noindent Clearly, if $\delta_i = 0$, this regret bound matches UCB for stochastic bandits. If $\delta_i  \neq 0$, we can also see that, acting using UCB resulting into very similar regret notation except that $\beta_i= c^2$ (see Appendix). Since $c^2(1-\delta_i)^2 \leq c^2$, we see that UA-UCB, by considering the uncertainty information in selection, produces a tighter bound. Note that when $\delta_i > 1 - \sqrt{\frac{1}{2c^2}}$, both UCB and UA-UCB lead to linear regret, i.e., the reward observed by agent is too much misleading w.r.t the real environment such that optimistic strategies based on estimation of $\hat{p}$ become not beneficial.   
}

\section{Uncertainty Adapted Monte Carlo Tree Search (UA-MCTS)}
%We first show that the performance of standard MCTS degrades significantly when $\hat{M}$ is imperfect. Figures \ref{fig:spc1}-\ref{fig:brk1} compare baseline MCTS using the perfect model $M$ (dashed green line), against MCTS using the imperfect model $\hat{M}$ (dashed red line). 

How can search in an imperfect model $\hat{M}$ improve decision-making in $M$? Of course, the optimal strategy for the agent is to exploit the mathematical relation between $\hat{M}$ and $M$; however, this information is unknown for the agent. In UA-MCTS, we adapt MCTS to behave more conservatively in states where the estimated uncertainty is larger. By doing so, we discourage, but do not completely give up on, searching through the more uncertain parts of the model.

%One design goal is to prevent the inaccuracies from compounding as the search tree grows. 
%In our implementation, the uncertainty estimate $\hat{U}$ is learned gradually while the agent is interacting with the environment. 

Guided by the insights from the corrupted bandit case, in UA-MCTS, we use four custom functions for \textit{selection}, \textit{expansion}, \textit{simulation}, and \textit{backpropagation} within an MCTS framework which utilizes this learned transition uncertainty. These four functions are described in subsections \ref{subsec:sel} - \ref{subsec:back}. 

\subsection{UA-Selection} \label{subsec:sel}
Similar to UA-UCB, in UA-Select, the exploration term in the standard UCT formula is dampened by a multiplicative term $1 - \alpha_i$, so that children with higher uncertainty will be explored less. $\alpha_i$ is a softmax of $\hat{U}$ with a temperature parameter set to the uncertainty factor $\tau > 0$. 
A lower $\tau$ makes the algorithm more sensitive to the predicted uncertainty. 
With a higher $\tau$ UA-Selection behaves more like the baseline MCTS selection.
%the new term $\alpha_i$ has higher values when the child node $v_i$ has a lower uncertainty and vice versa. 
Algorithm \ref{alg:uselection} describes the modified \textit{selection} function in pseudo-code, with the differences between UA-MCTS and the baseline UCT algorithm highlighted in red.

% prove added here
We prove that with this modification of UCT, our UA-MCTS remains complete in the sense that all nodes will be visited continuously in the limit as $N_I$ goes to infinity. We prove this by induction.
%rewrite in theorem format 
\begin{lemma}\label{lemma}
$\forall v \in Tree$, if $\Lim{N_I\to\infty} N(v)=\infty \Rightarrow \forall v_i \in Ch(v) \Lim{N_I\to\infty} N(v_i)=\infty$.\\
(Proof in Appendix) 
\end{lemma}

\begin{theorem}
(Completeness) If $ N_I \to \infty$, then $\forall v \in Tree, N(v) \to \infty$
\end{theorem}
\begin{proof}\renewcommand{\qedsymbol}{}
We prove this by induction on depth of the nodes in the tree.
\begin{enumerate}
    \item Induction base: $N(root)\to\infty$.
    \item Induction step:  $\Lim{N_I\to\infty} N(v)=\infty.  \Rightarrow \forall v_i\in Ch(v), \Lim{N_I\to\infty} N(v_i)=\infty$ (Proved in Lemma \ref{lemma}).
\end{enumerate}
\end{proof}

\begin{algorithm}[tb]
\textbf{~~~~Parameter}: temperature $\tau$ for softmax
\caption{Uncertainty Adapted Selection Algorithm.}\label{alg:uselection}
\begin{algorithmic}
\Function{Select}{$v$}
    \While{$v$ is expanded}
    \color{red}
    \For{$v_i \in Ch(v)$} 
        \State $\alpha_i \gets \frac{e^{\hat{U}(v_i) / \tau}}{\sum\limits_{v_j\in Ch(v)}e^{\hat{U}(v_j) / \tau}} $
    \EndFor
    \color{black}
    \State $v$ $\gets$ $\underset{v_i\in Ch(v)}{\mathrm{arg max}}\, \frac{Q(v_i)}{N(v_i)}+c\sqrt{\frac{\ln{N(v)}}{N(v_i)}} \cdot$  \color{red} (1 - $\alpha_i$) \color{black}
    \EndWhile   
    \State \Return $v$
\EndFunction
\end{algorithmic}
\end{algorithm}
\subsection{UA-Expansion} \label{subsec:exp}
%bias variance trade-off
To discourage searching parts of the tree where the model is more inaccurate, UA-Expansion gives children with higher uncertainty a lower chance to be added to the tree. 
While expanding node $v$, we initialize the uncertainty of each child $v_i$ based on the uncertainty of the transition $(S(v),a_i)$ that yielded $v_i$ and store it as $\hat{U}(v_i)$. This stored value can then be accessed by other parts of the UA-MCTS algorithm. 
Next, with probability $1-\tau/10$ we eliminate exactly one child from the tree. The choice of which child to eliminate is dependent on probabilities that are proportional to each child node's uncertainty.
The uncertainty factor $\tau$ is used in UA-Expansion as the elimination probability $\tau/10$; the scaling constant $1/10$ was chosen empirically.
Algorithm \ref{alg:uexp} shows the pseudo-code for UA-Expansion, with the modified parts in red. 

\begin{algorithm}[tp]
\caption{Uncertainty Adapted Expansion Algorithm}\label{alg:uexp}
\textbf{~~~~Parameter}: probability $\tau$ for deleting a node
\begin{algorithmic}
\Function{Expand}{$v$}
    \For{$a_i \in \mathcal{A}$}
        \State $s_i, r_i \gets \hat{M}(S(v),a_i)$
        \State create a node $v_i$ with state $s_i$ and reward $r_i$
        \color{red}
        \State $\hat{U}(v_i) \gets \hat{U}(S(v),a_i)$ 
        \color{black}
        \State $N(v_i) \gets 0$
        \State $Q(v_i) \gets 0$
    \EndFor
    \color{red}
    % \State $x \gets$ random number $\in (0, 1)$
    \State $x \sim Uniform(0, 1)$
    \If{$\sum\limits_{v_j\in Ch(v)} \hat{U}(v_j)> 0$ and $x < (1 - \tau/10)$}
        \For{$a_i \in \mathcal{A}$}
            \State $\alpha_i \gets \frac{\hat{U}(v_i)}{\sum\limits_{v_j\in Ch(v)} \hat{U}(v_j)}$
        \EndFor
        \State choose node $v_{i}$  with probability $\alpha_i$
        \State delete node $v_{i}$
    \EndIf
    \color{black}
    \State \Return a random child of $v$
\EndFunction
\end{algorithmic}
\end{algorithm}

\subsection{UA-Simulation} \label{subsec:sim}

In regular MCTS simulation, the returned value is the average of the $N_S$ rollouts.
To adapt simulation to transition uncertainty, we weigh each result based on a measure of the uncertainty of the whole rollout trajectory.
Assume $T_i$ is the rollout trajectory that the $i$th rollout took and $\mathcal{T}$ is the set of rollout trajectories. We define the uncertainty $\sigma(T_i)$ of trajectory $T_i=(s_1, a_1, s_2, a_2, \cdots, s_h, a_h, s_{h+1})$ as
\begin{equation*}\label{eq:sim2-1}
\sigma(T_i)\doteq\sum\limits_{k=1}^h \gamma^{k-1} \hat{U}(s_k, a_k)
\end{equation*}

The weight $\alpha_i$ for rollout $i$ is defined as the softmax

\begin{equation*}\label{eq:usim2-2}
\alpha_i \doteq \frac{e^{{-\sigma}(T_i) / \tau}}{\sum\limits_{j=1}^{N_S}e^{{-\sigma}(T_j) / \tau}}
\end{equation*}

Again, a lower $\tau$ makes $\alpha_i$ more sensitive to the predicted uncertainties. 
Let $g_i$ be the sum of discounted rewards for rollout $i$.
Then the weighted average return $G$ of all $N_S$ simulations is

\begin{equation*}\label{eq:usim2-3}
G \doteq\sum\limits_{i=1}^{N_S} \alpha_i\cdot g_i
\end{equation*}
Pseudo-code for UA-Simulation is presented in Appendix B.
% Algorithm \ref{alg:usim2} presents the pseudo-code for UA-Simulation, with the modified parts in red.

\subsection{UA-Backpropagation} \label{subsec:back}
% Conventional backpropagation looks at all the nodes in the same way. For instance, if a node has two children and it gets different values from each of them, after backpropagating from both of the children, the value of the parent node would be the average of both of the children. Over time if one of these children gets visited more then the value of the parent node gets closer to the value of the more visited child. More specifically, with normal backpropagation, the value of each node is the weighted average of the value of its children based on their visit counts. 

In Algorithm \ref{alg:ubackpropagate} UA-Backpropagation, nodes with more certainty have more impact on their parents. To do that, a child's value is modified with a multiplicative term that is based on its uncertainty before it is used to update its parent's value. 
This multiplicative term is the softmax of $-\hat{U}$ with a temperature parameter set to the uncertainty factor $\tau$, to assign children with lower uncertainty a higher weight.

\begin{algorithm}[tb]
\caption{Uncertainty Adapted Backpropagation Algorithm.}\label{alg:ubackpropagate}
\textbf{~~~~Parameter}: uncertainty factor $\tau$
\begin{algorithmic}
\Function{Backpropagate}{$v$, $value$}
    \While{$v$ is not NULL}
    \State $N(v) \gets N(v) + 1$
    \color{red}
    \State $\alpha = \frac{e^ {-\hat{U}(v) / \tau}}{\sum\limits_{n \in Ch(Par(v))} e^ {-\hat{U}(n) / \tau}} $
    \color{black}
    \State $Q(v) \gets Q(v) +$  \color{red} $\alpha$  \color{black} $\cdot value$
    \color{black}
    \State $value \gets value \cdot \gamma$
    \State $v \gets$ $Par(v)$ 
    \EndWhile   
\EndFunction
\end{algorithmic}
\end{algorithm}

\section{Experiments}\label{experiment}
We experimentally test the performance of UA-MCTS. In each test, we define a specific transition uncertainty $U$ and a method for learning its approximation $\hat{U}$.
We compare UA-MCTS with baseline MCTS in three modified MinAtar environments \citep{minAtar}. For each of the deterministic MinAtar environments -- Space Invaders, Freeway, and Breakout -- we briefly explain the modified domain, and discuss the experimental results. 
Also, we compare learning to improve the transition function with learning the uncertainty model in a toy environment setting.
Moreover, in Appendix A, we discuss the scenario where all paths to the goal contain uncertainty.

\subsection{Defining Uncertainty and Learning its Estimate} 
\label{subsec:learn-u}
The uncertainty estimate $U$ and the implementation of its approximation $\hat{U}$ can be chosen freely in our framework. For our experiments we use the following approach: 
Given $M$ and $\hat{M}$, we define the \textit{transition uncertainty} $U$ as the squared difference of the state vectors
\begin{equation*}
U(s, a)= \big( \hat{M}(s, a) - M(s, a) \big) ^2 .
\end{equation*}

A neural network uncertainty model $\hat{U}$ which approximates $U$ is learned by regression. After each action in the environment $M$, $U(s,a)$ is computed and the tuple $\langle s, a, U(s, a)\rangle$ is added to a buffer $B$. After every $I$ steps in the environment, the neural network $\hat{U}$ is trained for $E$ training steps.
$\tau$ decays exponentially with the update formula $\tau = \tau / 10 $ at each $I$ steps in the environment. 
This scheduling scheme for $\tau$ enables UA-MCTS to be more sensitive to the predicted uncertainty. 
Each training step performs a gradient descent update with a randomly selected batch $B' \subset B$ and loss function $L$:

\begin{equation*}
    \label{mse_loss}
    L = \sum_{s, a, U(s,a) \in B'} \bigg(\hat{U}(s, a) - U(s, a)\bigg) ^2
\end{equation*}

For each real world action decision, UA-MCTS is run with the current state as a start state for $N_I$ iterations. 
Whenever a node $v$ is added to the tree from transition $(s,a)$, we define its uncertainty as $\hat{U}(v) = \hat{U}(s,a)$. 
% For stability, we keep the uncertainty model $\hat{U}$ constant \textbf{how long? If we train after every I step we can't make it constant for ever so you mean it's constant in the I steps in the middle}. 
In the offline case, $\hat{U}=U$; in the online case, $\hat{U}$ is learned as above.

\subsection{Setup of Experiments}
We test our method on the three deterministic games Space Invaders, Freeway and Breakout in the MinAtar framework \citep{minAtar}. While the original games are all continuing tasks, we used an episodic version for our experiments.

We modify each game to be ``slightly broken'' by changing the transition function $M$.
The agent still believes it is playing the original game and plans using that model $\hat{M}$, which is no longer correct for the modified game.
We compare UA-MCTS and its four components with two baseline MCTS algorithms in this setting.
For each game, we study two uncertainty scenarios and compare a total of seven search algorithms:

1) In the \textit{offline scenario}, the agent has access to the true uncertainty $U$ of the model, but not to the perfect model $M$ itself. This scenario evaluates the performance of UA-MCTS with a ``perfect uncertainty'' model. The reason for this experiment is to separate out the difficulty of not knowing $M$ from the extra difficulty due to the training errors in the estimate $\hat{U}$.

2) In the \textit{online scenario}, the agent does not have access to $U$, and therefore has to learn an approximation $\hat{U}$ online from real experience. The agent starts with a random initialization of the uncertainty network and collects the buffer of transitions $B$ that is used to train $\hat{U}$ as explained in Section \ref{subsec:learn-u}.

The seven search algorithms tested are as follows:
the two baselines use standard MCTS without any UA modifications. 

1) ``True Model'' MCTS is allowed to use $M$ in its search. It is shown as a horizontal dashed green line in the figures.

2) ``Corrupted Model'' MCTS uses $\hat{M}$ for all its planning, and is shown as a horizontal dashed red line in the figures.

3-7) All versions of UA-MCTS work as follows: they use $\hat{M}$ for all their planning. In the offline scenario, they use the true $U$ from the beginning, as explained above. In the online scenario, they start by running the Corrupted Model MCTS (no UA modifications), and gain real experience from the moves played. 
The UA-MCTS versions labeled ``Backpropagation'', ``Selection'', ``Expansion''  and ``Simulation'' use the UA modifications only for this one component of MCTS, and use unmodified MCTS for the other three parts. The ``Combined'' version uses all four enhancements.

In the offline scenario we show results for all seven algorithms.
In the online scenario, for simplicity we only show results for the combined version of UA-MCTS and the two MCTS baselines. 

% For each combination of game and algorithm we performed a parameter sweep over the uncertainty factor $\tau$ from the set $\{0.1, 0.5, 0.9\}$ and for the exploration constant $c$ from the set $\{0.5, 1, \sqrt{2}, 2\}$.
For each combination of game and algorithm we performed a parameter sweep over the exploration constant $c$ from the set $\{0.5, 1, \sqrt{2}, 2\}$.
For the ``Combined'' version in the online scenario, we used the best $c$ found for the offline scenario. In the offline scenario the uncertainty factor $\tau$ is set to $0.1$ (a small number so that UA-MCTS is more sensitive to the true uncertainty), and in the online scenario the $\tau$ is initialized to $10$, then decays until it reaches $0.1$. 
The uncertainty model in the online scenario is a fully connected neural network with two hidden layers of 128 hidden units each. The number of training steps $E$ and training frequency $I$ are both 5000, and the step size is $10^{-3}$ for the Adam optimizer \citep{adam}.
Table \ref{ptable} shows a list of other hyperparameters used in UA-MCTS.

% and no knowledge of any changes made to the environment, i.e. it uses the unmodified transition function.

\subsection{Space Invaders}
\begin{figure}[htb]
    \centering
    \includegraphics[width=7cm]{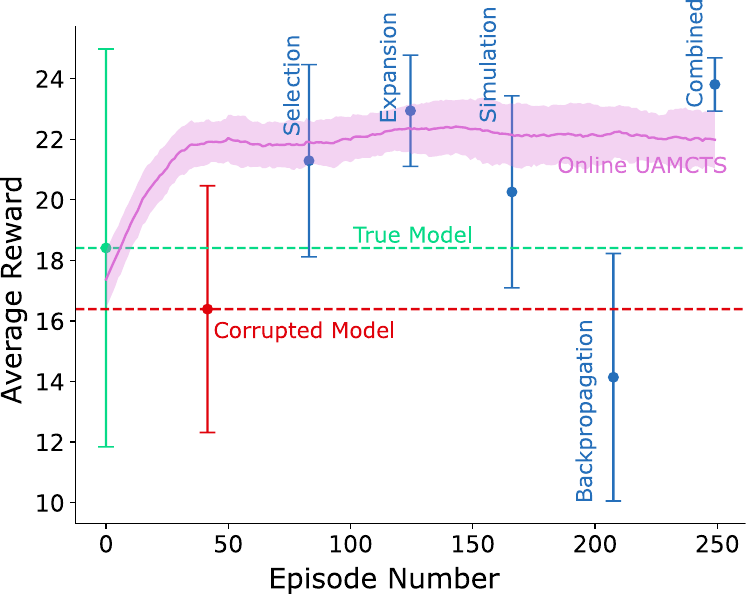}
    \caption{
    % The comparison of UA-MCTS and its components (vertical bars) with MCTS baselines in Space Invaders environment for online and offline scenarios.
    Offline (vertical bars) and online scenarios for Space Invaders. Bars and shaded areas show $mean \pm std$ of rewards over 100 runs for the offline and 15 for the online scenarios. For the corrupted and true models (two-leftmost cases), the best $c$ is $2$ and $\sqrt{2}$ respectively. The best $c$ parameter chosen for each of the UA-MCTS algorithms (the remaining five cases) are [$2$, $2$, $1$, $0.5$, $\sqrt{2}$] from left to right.
    For the online scenario, we plot the moving average of the reward with a window size of 50.
    }
    \label{fig:spc1}
\end{figure}

% \todo{for all three experiments: 1. check what the error bars are actually showing, or replace by box plot.(added the error bar explanation on the figure captions) 2. add numbers when discussing performance. or differences in perf. 3. Add discussion of ablation results and overall results, and possible explanations, conjectures}
In Space Invaders, the agent tries to eliminate 24 enemies by shooting at them \citep{minAtar}.
We modified this environment by disabling the shoot action in five out of ten positions, at indices 2, 3, 4, 5, and 6.
As shown in Figure \ref{fig:spc1}, the performance of the baseline MCTS drops from an average reward of 18.4 for the True Model MCTS to 16.3 for the Corrupted Model MCTS, where the agent is unaware of its limitations on shooting.

In the offline scenario,
UA-MCTS worked surprisingly well, and even exceeded the performance of the True Model MCTS in all configurations except UA-backpropagate-only.
As a single modification, UA-expand-only performed best.
The combined UA-MCTS achieved an average reward of 23.8, which is close to the perfect play reward of 24. Even in the online setting, after training $\hat{U}$, combined UA-MCTS outperforms the True Model MCTS. 
In this game, low value states are strongly correlated with high uncertainty states where the agent cannot shoot. Since UA-MCTS discourages visits to these states, its search becomes more efficient than the True Model MCTS despite the imperfect model.
%but it is likely limited by the relatively low search budget ($N_I=10$).

\subsection{Freeway}
\begin{figure}[htb]
    \centering
    \includegraphics[width=7cm]{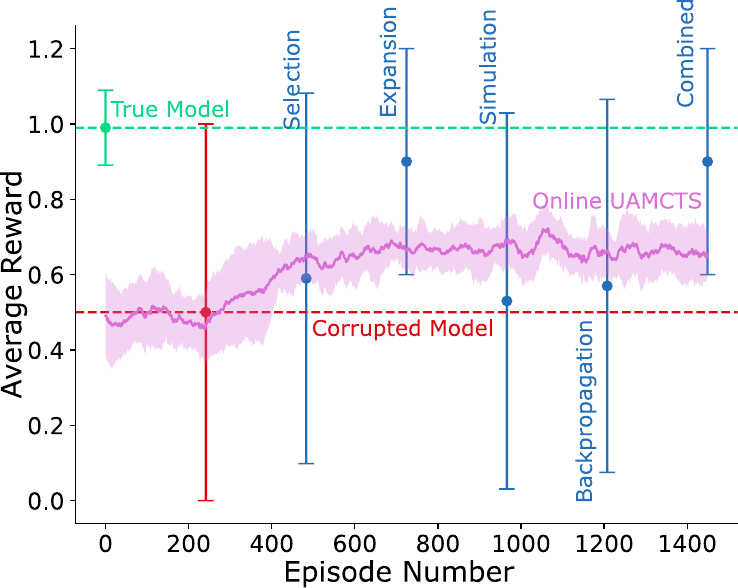}
    \caption{
    Offline (vertical bars) and online scenarios for Freeway. Bars and shaded areas show $mean \pm std$ of rewards over 100 runs for the offline and 15 for the online scenarios. For the corrupted and true models (two-leftmost cases), the best $c$ is $0.5$ and $2$ respectively. The best $c$ parameter chosen for each of the UA-MCTS algorithms (the remaining five cases) are [$2$, $2$, $\sqrt{2}$, $2$, $\sqrt{2}$] from left to right.
    For the online scenario, we plot the moving average of the reward with a window size of 50.}
    \label{fig:frw1}
\end{figure}

In Freeway, the agent's goal is to reach the top of the screen without touching enemies along the way \citep{minAtar}. 
In the modified game $M$, executing the action ``None'' moves the agent up in six out of ten locations. 
Figure \ref{fig:frw1} shows that offline, combined UA-MCTS achieves the average reward of 0.9 and outperforms Corrupted Model MCTS. The individual performance of each component is better than the Corrupted Model, with UA-expand-only again performing best.
In the online setting, UA-MCTS is able to outperform the Corrupted Model MCTS, but cannot approach the True Model MCTS. Comparing the two results (offline and online) for this game, we conjecture that the uncertainty network's inaccuracies limit the performance. 

The online UA-MCTS agent converges to its best performance in Space Invaders using fewer episodes than in Freeway. The reason is that Space Invaders episodes are longer, so the agent can learn from more environment interaction samples.

% With the perfect uncertainty model \ref{fig:frw1}, we can observe that our agent outperforms the original MCTS. The search tree is changed in a way that it avoids expanding the tree for the None action in the modified positions as it is uncertain about this action. It tries to find a path to the goal using other actions (Up, Down) in the modified positions.

\begin{table}[t]
\centering
\renewcommand{\arraystretch}{1}
\renewcommand{\tabcolsep}{0.5cm}
\scalebox{0.8}{%
\begin{tabular}{c||c c c}                   & Space Invaders & Freeway  & Breakout                \\ \hline\hline
$N_I$   & 10    & 100    & 100  \\ 
$D_S$   & 20     & 50   & 50 \\ 
$N_S$   & 10    & 10    & 10  \\ 
\end{tabular}
}
\caption{Experiment settings}
\label{ptable}
\end{table}

\subsection{Breakout}
\begin{figure}[htb]
    \centering
    \includegraphics[width=7cm]{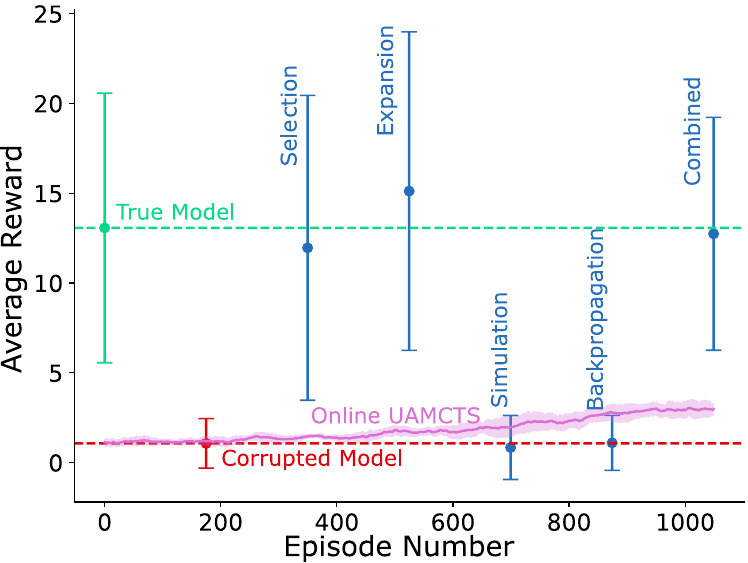}
    \caption{Offline (vertical bars) and online scenarios for Breakout. Bars and shaded areas show $mean \pm std$ of rewards over 100 runs for the offline and 15 for the online scenarios. For the corrupted and true models (two-leftmost cases), the best $c$ is 2 and $\sqrt{2}$ respectively. The best $c$ parameter chosen for each of the UA-MCTS algorithms (the remaining five cases) are [$1$, $2$, $\sqrt{2}$, $2$, $\sqrt{2}$] from left to right.
    For the online scenario, we plot the moving average of the reward with a window size of 50.}
    \label{fig:brk1}
\end{figure}

In Breakout, the agent controls a paddle at the bottom of the screen, which is used to bounce a ball back up. The goal is to hit and destroy as many bricks as possible \citep{minAtar}. 
In our modified game, the paddle fails to stop the ball in two out of ten positions, ending the game. To overcome this, the agent must plan ahead to keep the ball in play without using these corrupted positions.
Figure \ref{fig:brk1} shows that UA-MCTS outperformed the Corrupted Model MCTS baseline and performed almost as well as the True Model MCTS.
UA-select-only and UA-expand-only also performed very well, while UA-backpropagate-only and UA-simulate-only performed poorly.
%""
%Figure \ref{fig:brk1} shows that offline, UA-expand-only and UA-select-only work well. Again, combined UA-MCTS is best, but it is not perfect.
%There is a significant performance drop in the online setting, with only a small improvement on Corrupted Model MCTS. This game shows some limitations in our current implementation. It seems that the agent did not have the training samples which allow it to accurately learn the uncertainty. We looked into the buffer $B$; it has many repetitive samples.
%""
% Again, combined UA-MCTS is best, but it is not perfect.
% Moreover, the offline UA-MCTS outperformed the Corrupted Model MCTS.
% , but not UA-expand-only.
% We can say the UA-MCTS is not always better than each individual component, but it performs better than Corrupted Model MCTS in all the three environments.
Another observation is the performance drop in the online setting w.r.t the offline setting, although it still performed better than the Corrupted Model MCTS. Since the main difference between the two settings is the uncertainty, the online setting is likely limited by its learned uncertainty's lower accuracy. The effect of this inaccuracy, and potential improvements, remain an interesting future direction of research.
% This game shows some limitations in our current implementation. It seems that the agent did not have the training samples which allow it to accurately learn the uncertainty. 
% We looked into the buffer $B$; it has many repetitive samples.
% For Breakout, the uncertainty MSE ($3.84 \times 10^{-5}$) is about twice that of Space Invaders and Freeway. 
% In Space Invaders and Freeway the uncertainty MSE gets lower than ($1.0 \times 10^{-7}$), but it cannot get lower than ($1.0 \times 10^{-6}$) in Breakout.

\subsection{Learning the Uncertainty VS. Learning the Transition Function}\label{sec:lrn-trsn}
\begin{figure}[tb]
    \centering
    \includegraphics[scale=0.2]{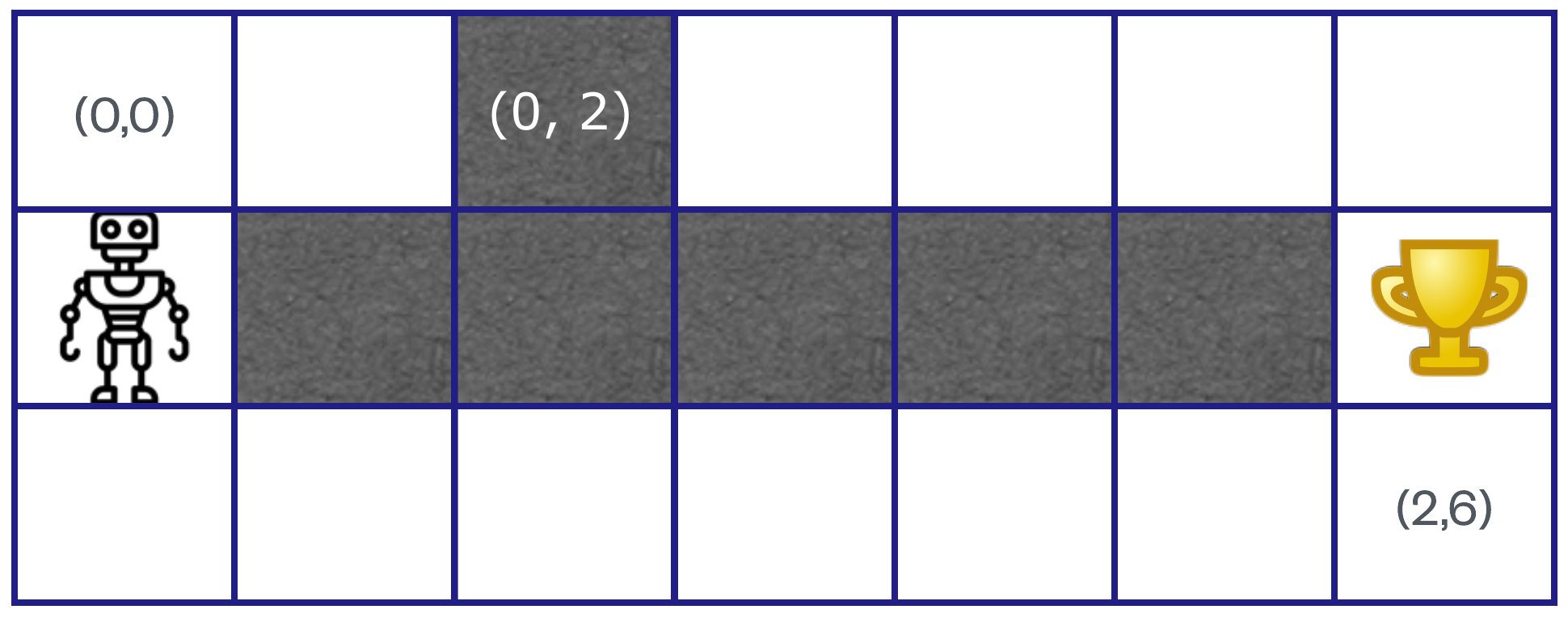}
    \caption{2-Way GridWorld Environment. White cells are empty and grey cells are walls.}
    \label{fig:tw_snapshot}
\end{figure}

An alternative approach to learning the uncertainty, then applying that to UA-MCTS, is to use MCTS with a learned transition function that approximates the true environment. 
We justify the choice of focusing on the former in this paper as follows.
We believe that training the transition function is generally more difficult than training the uncertainty because state representations tend to be more complex.
In contrast, uncertainty can be represented with a single scalar value regardless of the environment.
As an example, the feature vector representing each state contains 306, 34, and 107 scalar elements for Space Invaders, Freeway, and Breakout, respectively.
% in the order of 100 \textbf{exact values; maybe saying 306, Y, Z for each of the mentioned envs} floating point numbers.
Therefore, with the same amount of resources, we should be able to learn a more accurate uncertainty model than a transition model. 
% When combined with UA-MCTS, this uncertainty model should outperform MCTS with a transition function that uses roughly the same amount of resources.
\cite{cmax, cmax++} also mention the same intuition, which motivated their approach.
To illustrate this with an example, we designed a toy environment called the 2-Way GridWorld. 
A snapshot of this environment is presented in Figure \ref{fig:tw_snapshot}. The agent starts at position $(1,0)$ and at each time step it can choose any of the four actions: ``Up'', ``Down'', ``Left'', and ``Right''. The terminal state is at position $(1,6)$. When the agent reaches the position $(1,6)$, the game terminates and the agent gets a reward of $10$. The maximum number of steps in each episode is $50$. In the corrupted model the agent does not know that the wall in position $(0,2)$ exists. Therefore, it assumes that it can reach the goal either from the top path or from the bottom path. The reality is that the top path is not a valid plan. 

We compare the two scenarios:
1) Using combined UA-MCTS and learning the uncertainty online: we use combined UA-MCTS and the same method and setup for learning the uncertainty as in the MinAtar environments, except that the uncertainty model is a fully connected neural network with no hidden layers. $\tau$ is initialized to $10$ and decays until it reaches $0.1$.
2) Using MCTS and learning the transition function online: We use MCTS and learn the transition function online. The process of learning the transition function is the same as learning the uncertainty model, except that the buffer $B$ is filled with $\langle s, a, M(s, a)\rangle$ samples. For the first $I$ steps, the corrupted model is used to gather a buffer, since the transition function has not been trained yet.

\begin{figure}[tb]
    \centering
    \includegraphics[width=7cm]{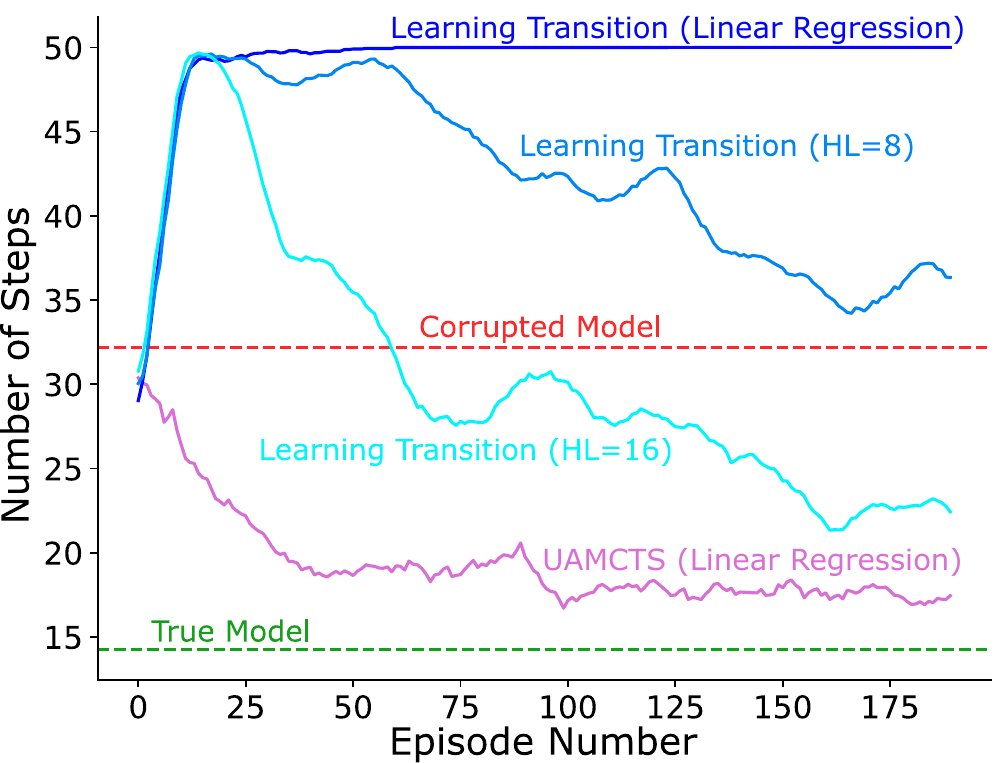}
    \caption{Comparison between learning the transition function with different hidden layer (HL) sizes and UA-MCTS method.
    The average number of steps to the goal is over 30 runs. The exploration parameter $c$ for ``True Model'' and ``Corrupted Model' is $2$ and $\sqrt{2}$ respectively. For the other scenarios $c$ is $\sqrt{2}$.}
    \label{fig:tw}
\end{figure}

We used three different transition functions: a) a linear regression, b) a fully connected neural network with 1 hidden layer consisting of 8 units, and c) a fully connected neural network with 1 hidden layer with 16 units. All models predict the feature vector of the next state.
We also compare with the two baselines ``True Model'' and ``Corrupted Model''.
We performed a parameter sweep over the exploration constant $c$ from the set $\{0.5, 1, \sqrt{2}, 2\}$ for the ``True Model'' and ``Corrupted Model''. We used the best $c$ found for the ``Corrupted Model'' for the two scenarios.
In this set of experiments $N_I=10$, $N_S=10$, $D_S=30$, $I=300$, and $E=5000$.
Figure \ref{fig:tw} presents the result of this experiment. We can observe that the combined UA-MCTS converges to a result much better than the Corrupted Model MCTS and close to the True Model MCTS. However, when trying to learn the transition function using the same network capacity, the agents could not find a path to the goal at all. The agent improves as the capacity of the transition function increases, but is still not able to outperform UA-MCTS with 16 units in its hidden layer.
%This experiment shows that learning uncertainty is an easier task than learning the transition function.
%agent not aware!

\section{Related Work}
% \textbf{which type of uncertainty? I'm not sure what to answer for this question?}
There are numerous techniques which quantify and use uncertainty for MBRL.
\cite{safeRL} captured uncertainty using ensembles of LSTM and Monte Carlo dropout, and changed the behaviour of their agent to act more cautiously in the uncertain parts by introducing a cost function for model predictive control (MPC), which can be considered a search algorithm.

Several approaches used uncertainty, but not as a component of an explicit search.
\cite{zaheer,AdaMVE,steve} modified the model value expansion (MVE) algorithm by taking model uncertainty into consideration. 
\cite{zaheer} and \cite{AdaMVE} selected the bootstrapping depth using model uncertainty.
\cite{hallucinating} investigated the effect of model updates in both forward and backward directions with an imperfect model. 
\cite{BMPO} used forward and backward models in model-based policy optimization \citep{MBPO} in order to reduce accumulative model error while maintaining a similar update depth.
\cite{self-correcting-models} designed a way to learn the model which reduces accumulative model error in deep lookahead.

\textsc{Cmax} and \textsc{Cmax++} \citep{cmax,cmax++} are robust online path-planning algorithms in a real-time A* search framework, which can work around imperfect model predictions. As in our work, imperfect states are identified by comparing against the real environment during interactions. At runtime, \textsc{Cmax} completely disables parts of the model that are found to be in conflict. Examples of this include malfunctioning motors or overloaded robotic arms. To find a solution, at least one perfectly modelled path to the goal needs to exist. In contrast, UA-MCTS does not hard-prune such states, but shapes the search to prefer states with lower estimated uncertainty.

\section{Conclusion and Future Work}
UA-MCTS improves MCTS by learning a simple uncertainty model, instead of trying to learn corrections to a complex transition function. The method modifies the four main components of MCTS in a ``risk-averse'' way so that the search is directed away from the more uncertain parts of the model. To test UA-MCTS in practice, we developed a definition of uncertainty for a fixed given model, and a learned approximation that works with UA-MCTS. We tested UA-MCTS on the MinAtar testbed. Our results show that: 1) baseline MCTS suffers from severe performance degradation in the face of model uncertainty. 2) UA-MCTS outperforms MCTS for imperfect models. It can recover from performance degradation, or at least lessen its effects when used in conjunction with a learned uncertainty estimate. 3) The precision of the learned uncertainty model used by UA-MCTS has a very strong effect on agent performance. Improving this estimate requires further investigation. Other future directions include: investigating the impact of varying degrees of model imperfection on both UA-MCTS and baseline MCTS, especially the ``risk-seeking'' case where high uncertainty states must be explored, and comparing different uncertainty estimation techniques for UA-MCTS.

\bibliography{aaai24}

\begin{thebibliography}{26}
\providecommand{\natexlab}[1]{#1}

\bibitem[{Abbas et~al.(2020)Abbas, Sokota, Talvitie, and White}]{zaheer}
Abbas, Z.; Sokota, S.; Talvitie, E.; and White, M. 2020.
\newblock Selective {D}yna-style planning under limited model capacity.
\newblock In \emph{International Conference on Machine Learning}, 1--10. PMLR.

\bibitem[{Arneson, Hayward, and Henderson(2010)}]{MoHex}
Arneson, B.; Hayward, R.; and Henderson, P. 2010.
\newblock {M}onte {C}arlo tree search in {H}ex.
\newblock \emph{IEEE Transactions on Computational Intelligence and AI in
  Games}, 2(4): 251--258.

\bibitem[{Banerjee(2020)}]{MIMHex}
Banerjee, D. 2020.
\newblock HEX and Neurodynamic Programming.
\newblock \emph{arXiv preprint arXiv:2008.06359}.

\bibitem[{Browne et~al.(2012)Browne, Powley, Whitehouse, Lucas, Cowling,
  Rohlfshagen, Tavener, Perez, Samothrakis, and Colton}]{MCTSSurvey}
Browne, C.~B.; Powley, E.; Whitehouse, D.; Lucas, S.~M.; Cowling, P.~I.;
  Rohlfshagen, P.; Tavener, S.; Perez, D.; Samothrakis, S.; and Colton, S.
  2012.
\newblock A survey of {M}onte {C}arlo tree search methods.
\newblock \emph{IEEE Transactions on Computational Intelligence and AI in
  games}, 4(1): 1--43.

\bibitem[{Buckman et~al.(2018)Buckman, Hafner, Tucker, Brevdo, and Lee}]{steve}
Buckman, J.; Hafner, D.; Tucker, G.; Brevdo, E.; and Lee, H. 2018.
\newblock Sample-Efficient Reinforcement Learning with Stochastic Ensemble
  Value Expansion.
\newblock In \emph{Proceedings of the 32nd International Conference on Neural
  Information Processing Systems}, NIPS'18, 8234–8244. Curran Associates Inc.

\bibitem[{Cazenave(2010)}]{MEG}
Cazenave, T. 2010.
\newblock Nested {M}onte-{C}arlo expression discovery.
\newblock In \emph{ECAI 2010}, 1057--1058. IOS Press.

\bibitem[{Jafferjee et~al.(2020)Jafferjee, Imani, Talvitie, White, and
  Bowling}]{hallucinating}
Jafferjee, T.; Imani, E.; Talvitie, E.; White, M.; and Bowling, M. 2020.
\newblock Hallucinating value: A pitfall of {D}yna-style planning with
  imperfect environment models.
\newblock \emph{arXiv preprint arXiv:2006.04363}.

\bibitem[{Janner et~al.(2019)Janner, Fu, Zhang, and Levine}]{MBPO}
Janner, M.; Fu, J.; Zhang, M.; and Levine, S. 2019.
\newblock When to Trust Your Model: Model-Based Policy Optimization.
\newblock \emph{Advances in Neural Information Processing Systems}, 32:
  12519--12530.

\bibitem[{Kingma and Ba(2014)}]{adam}
Kingma, D.~P.; and Ba, J. 2014.
\newblock Adam: A method for stochastic optimization.
\newblock \emph{arXiv preprint arXiv:1412.6980}.

\bibitem[{Kocsis, Szepesv{\'a}ri, and Willemson(2006)}]{uct}
Kocsis, L.; Szepesv{\'a}ri, C.; and Willemson, J. 2006.
\newblock Improved {M}onte-{C}arlo search.
\newblock \emph{Univ. Tartu, Estonia, Tech. Rep}, 1.

\bibitem[{Lai et~al.(2020)Lai, Shen, Zhang, and Yu}]{BMPO}
Lai, H.; Shen, J.; Zhang, W.; and Yu, Y. 2020.
\newblock Bidirectional Model-based Policy Optimization.
\newblock In \emph{International Conference on Machine Learning}, 5618--5627.
  PMLR.

\bibitem[{Lattimore and Szepesv{\'a}ri(2020)}]{lattimore2020bandit}
Lattimore, T.; and Szepesv{\'a}ri, C. 2020.
\newblock \emph{Bandit algorithms}.
\newblock Cambridge University Press.

\bibitem[{Lu et~al.(2016)Lu, Wang, Wang, and Wang}]{FA}
Lu, J.; Wang, X.; Wang, D.; and Wang, Y. 2016.
\newblock Parallel {M}onte {C}arlo tree search in perfect information game with
  chance.
\newblock In \emph{2016 Chinese Control and Decision Conference (CCDC)},
  5050--5053. IEEE.

\bibitem[{L{\"u}tjens, Everett, and How(2019)}]{safeRL}
L{\"u}tjens, B.; Everett, M.; and How, J.~P. 2019.
\newblock Safe reinforcement learning with model uncertainty estimates.
\newblock In \emph{2019 International Conference on Robotics and Automation
  (ICRA)}, 8662--8668. IEEE.

\bibitem[{Mansley, Weinstein, and Littman(2011)}]{HOOT}
Mansley, C.; Weinstein, A.; and Littman, M. 2011.
\newblock Sample-based planning for continuous action {M}arkov {D}ecision
  {P}rocesses.
\newblock In \emph{Twenty-First International Conference on Automated Planning
  and Scheduling}.

\bibitem[{Nijssen and Winands(2010)}]{checkers}
Nijssen, J.; and Winands, M. 2010.
\newblock Enhancements for multi-player {M}onte-{C}arlo tree search.
\newblock In \emph{International Conference on Computers and Games}, 238--249.
  Springer.

\bibitem[{Sabharwal, Samulowitz, and Reddy(2012)}]{MIP}
Sabharwal, A.; Samulowitz, H.; and Reddy, C. 2012.
\newblock Guiding combinatorial optimization with {UCT}.
\newblock In \emph{International conference on integration of artificial
  intelligence (AI) and operations research (OR) techniques in constraint
  programming}, 356--361. Springer.

\bibitem[{Saffidine(2008)}]{YOPT}
Saffidine, A. 2008.
\newblock Utilisation d{U}{C}{T} au {H}ex.
\newblock \emph{Ecole Normale Super. Lyon, France, Tech. Rep}.

\bibitem[{Silver et~al.(2016)Silver, Huang, Maddison, Guez, Sifre, Van
  Den~Driessche, Schrittwieser, Antonoglou, Panneershelvam, Lanctot
  et~al.}]{AlphaGo}
Silver, D.; Huang, A.; Maddison, C.~J.; Guez, A.; Sifre, L.; Van Den~Driessche,
  G.; Schrittwieser, J.; Antonoglou, I.; Panneershelvam, V.; Lanctot, M.;
  et~al. 2016.
\newblock Mastering the game of {G}o with deep neural networks and tree search.
\newblock \emph{nature}, 529(7587): 484--489.

\bibitem[{Silver, Sutton, and M{\"u}ller(2008)}]{silver2008sample}
Silver, D.; Sutton, R.~S.; and M{\"u}ller, M. 2008.
\newblock Sample-based learning and search with permanent and transient
  memories.
\newblock In \emph{Proceedings of the 25th International Conference on Machine
  Learning}, 968--975.

\bibitem[{Sutton and Barto(2018)}]{richbook}
Sutton, R.~S.; and Barto, A.~G. 2018.
\newblock \emph{Reinforcement learning: An introduction}.
\newblock MIT press.

\bibitem[{Talvitie(2017)}]{self-correcting-models}
Talvitie, E. 2017.
\newblock Self-correcting models for model-based Reinforcement Learning.
\newblock In \emph{Thirty-First AAAI Conference on Artificial Intelligence}.

\bibitem[{Vemula, Bagnell, and Likhachev(2021)}]{cmax++}
Vemula, A.; Bagnell, J.~A.; and Likhachev, M. 2021.
\newblock CMAX++: Leveraging Experience in Planning and Execution using
  Inaccurate Models.
\newblock In \emph{Proceedings of the AAAI Conference on Artificial
  Intelligence}, volume~35, 6147--6155.

\bibitem[{Vemula et~al.(2020)Vemula, Oza, Bagnell, and Likhachev}]{cmax}
Vemula, A.; Oza, Y.; Bagnell, J.~A.; and Likhachev, M. 2020.
\newblock Planning and Execution using Inaccurate Models with Provable
  Guarantees.
\newblock In \emph{Proceedings of Robotics: Science and Systems}.

\bibitem[{Xiao et~al.(2019)Xiao, Wu, Ma, Schuurmans, and M{\"u}ller}]{AdaMVE}
Xiao, C.; Wu, Y.; Ma, C.; Schuurmans, D.; and M{\"u}ller, M. 2019.
\newblock Learning to combat compounding-error in model-based reinforcement
  learning.
\newblock \emph{NeurIPS 2019 Deep Reinforcement Learning Workshop. arXiv
  preprint arXiv:1912.11206}.

\bibitem[{Young and Tian(2019)}]{minAtar}
Young, K.; and Tian, T. 2019.
\newblock Min{A}tar: An Atari-inspired testbed for thorough and reproducible
  {R}einforcement {L}earning experiments.
\newblock \emph{arXiv preprint arXiv:1903.03176}.

\end{thebibliography}

\end{document}

\onecolumn %% Turn this off if single column is desired for the supplement

\maketitle

%Appendix \ref{icy} shows an experiment on a toy problem in which the agent has to choose a path that includes uncertainty.
%Appendix \ref{simulation} shows the pseudo-code for the modified \textit{simulation} function described in Section 3.3 of the paper.

\appendix
\section{UA-MCTS with ``mandatory'' uncertain paths}\label{icy}
\begin{figure}[htb]
    \centering
    \includegraphics[width=5cm]{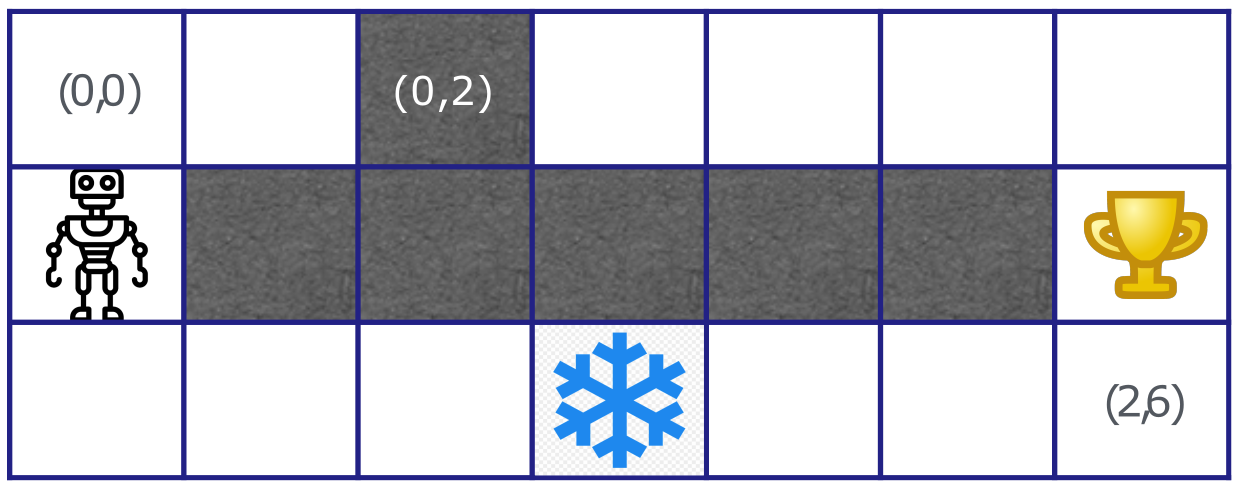}
    \caption{Icy 2-Way GridWorld Environment. White cells are empty and grey cells are walls. The blue cell is the icy one.}
    \label{fig:twicy_snapshot}
\end{figure}

\begin{figure}[htb]
    \centering
    \includegraphics[width=6cm]{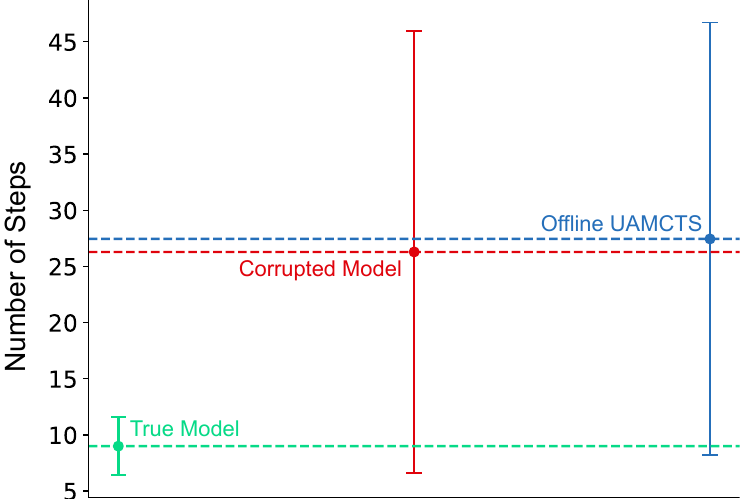}
    \caption{Result of the experiments in the Icy 2-Way GridWorld problem.
    Bars show $mean \pm std$ of rewards over 30 runs. 
    The exploration parameter $c$ for ``True Model'' and ``Corrupted Model' are both $0.5$. For UA-MCTS, $c$ is $1$.}
    \label{fig:twicy}
\end{figure}

While UA-MCTS discourages using uncertain parts of the model, it does not completely give up on them. If the paths to the goal contain uncertainty, UA-MCTS is still able to find them. We show this through an experiment. Assume the position $(2,3)$ in the 2-Way GridWorld problem is icy and slippery. We call this new environment Icy 2-Way GridWorld. A snapshot of this environment is presented in Figure \ref{fig:twicy_snapshot}. If the agent moves to this position, it slips and moves to the next position. However, the agent is not aware of the icy position and the wall in position $(0,2)$. Therefore, there is only one path to the goal and there is uncertainty in the path. If the agent wants to reach the goal, it must pass through the uncertain state. 
In this set of experiments $N_I=10$, $N_S=10$, $D_S=30$, and $\tau=10$. 
We performed a parameter sweep over the exploration constant $c$ from the set $\{0.5, 1, \sqrt{2}, 2\}$.
Figure \ref{fig:twicy} presents the performance of the ``True Model'' and ``Corrupted Model'' agents as well as offline UA-MCTS in this environment. We present the performance of the offline version to separate out the complications of learning the transition uncertainty.
UA-MCTS does find a path to the goal even though this path contains transition uncertainty.
This shows that UA-MCTS does not ignore parts of the model that include transition uncertainty.

% \section{Additional simulation results}
% Table~\ref{tab:supp-data} lists additional simulation results; see also \citet{einstein} for a comparison. 

% \begin{table}[!h]
%     \centering
%     \caption{An Interesting Table.} \label{tab:supp-data}
%     \begin{tabular}{rl}
%       \toprule % from booktabs package
%       \bfseries Dataset & \bfseries Result\\
%       \midrule % from booktabs package
%       Data1 & 0.12345\\
%       Data2 & 0.67890\\
%       Data3 & 0.54321\\
%       Data4 & 0.09876\\
%       \bottomrule % from booktabs package
%     \end{tabular}
% \end{table}

% \section{Math font exposition}
% % NOTE: necessary when ptmx or no mathfont class option is given
% \providecommand{\upGamma}{\Gamma}
% \providecommand{\uppi}{\pi}
% How math looks in equations is important:
% \begin{equation*}
%   F_{\alpha,\beta}^\eta(z) = \upGamma(\tfrac{3}{2}) \prod_{\ell=1}^\infty\eta \frac{z^\ell}{\ell} + \frac{1}{2\uppi}\int_{-\infty}^z\alpha \sum_{k=1}^\infty x^{\beta k}\mathrm{d}x.
% \end{equation*}
% However, one should not ignore how well math mixes with text:
% The frobble function \(f\) transforms zabbies \(z\) into yannies \(y\).
% It is a polynomial \(f(z)=\alpha z + \beta z^2\), where \(-n<\alpha<\beta/n\leq\gamma\), with \(\gamma\) a positive real number.

% \bibliography{uai2023-template}

\section{UA-Simulation}\label{simulation}

Algorithm \ref{alg:usim2} describes the modified \textit{simulation} function mentioned in Section 3.3 in pseudo-code, with the modified parts in red.

\begin{algorithm}[htb]
\textbf{~~~~Parameter}: uncertainty factor $\tau$
\caption{Uncertainty Adapted Simulation}\label{alg:usim2}
\begin{algorithmic}
\Function{Simulate}{$s, depth$}
    \For{$i \gets 1$ to $N_S$}
        \State $g_i,$
        \color{red}${\sigma}(T_i)$
        \color{black}$\gets
        \Call{Rollout}{s}$
        \State $\alpha_i \gets 1/N_S$
    \EndFor
    \color{red}
    \For{$i \gets 1$ to $N_S$}
    
        \State $\alpha_i \gets \frac{e^{{-\sigma}(T_i) / \tau}}{\sum\limits_{j=1}^{N_S}e^{{-\sigma}(T_j) / \tau}}$
    \EndFor
    \color{black}
    \State \Return $\sum\limits_{i=1}^{N_S}$ $\alpha_i$
    \color{black}$\cdot g_i$
\EndFunction
\newline
\Function{Rollout}{$s$}
    \State $count \gets 0$
    \State $rewards \gets 0$
    \color{red}
    \State $\sigma \gets 0$
    \color{black}
    \State $discount \gets 1$
    \While{$s$ is not terminal $\And count < D_S$ }
        \State choose a random action $a$ from $\mathcal{A}$
        \State $s, r \gets \hat{M}(s,a)$
        \State $count \gets count + 1$
        \State $rewards \gets rewards + discount \cdot r$
        \color{red}
        \State $\sigma \gets \sigma + discount \cdot \hat{U}(s, a)$
        \color{black}
        \State $discount \gets discount\cdot \gamma $
    \EndWhile
    \State \Return $rewards$, \color{red} $\sigma$ \color{black}
\EndFunction
\end{algorithmic}
\end{algorithm}

\section{Proof of UA-UCB Regret Bound}\label{regret}
\noindent Recall that, in a corrupted $K$-armed bandit, we have the following two models.    
\begin{enumerate}
    \item The observed model reward distribution: \\
    $\hat{p}_i , \hat{\mu}_i, : i \in \{1, ... , k\}$
    
    \item The real reward distribution: \\
    $p_i , \mu_i : i \in \{1, ... , k\}$
\end{enumerate}

\noindent The following score function is used for UA-UCB selection. 
\begin{equation}
    \mathit{UA-UCB}_i(t) = \hat{x}_{i,t-1} + c \sqrt{\dfrac{\ln{(t-1)}}{N_{i,t-1}}} \cdot (1-\delta_i)
\end{equation}

% \noindent $N_{i, t-1}$ denote the action $i$ has been take for $N$ times during the first $t-1$ tries.
\noindent $N_{i, t-1}$ denote the number of times action $i$ has been visited during the first $t-1$ tries.
$\hat{x}_{i,t-1}$ is the approximate of $\hat{\mu}_i$ at time $t-1$.
We can only sample from model $\hat{p}_i$, and we assume the absolute value of difference is noted as below: 
\begin{equation}
    \delta_i = |\hat{\mu}_i - \mu_i| \in [0,1]
\end{equation}
In reality, $\delta$ is usually unknown or can only be gradually revealed. To simplify the analysis and avoid extra assumption on how $\delta$ is revealed to the learning agent, we assume $\delta$ is known. A special case is that when $\delta_i=0$, UA-UCB becomes normal UCB, which is an expected result since in this case there is no difference between $p_i$ and $\hat{p}_i$. In below we show regret bound for UA-UCB for the general cases.    
%If $\delta_i=1$, UA-UCB only includes  the first term $\hat{x}_{i,t-1}$ and the regret will be linear.

\noindent Let $\hat{i}^{*}$ and $i^{*}$ be the optimal action for $\hat{p}_i$ and $p_i$, then 
\begin{IEEEeqnarray}{lcr}
    \hat{\Delta}_i &=& \hat{\mu}_{\hat{i}^*} - \hat{\mu}_{i} \\
    \Delta_i &=& \mu_{i^*} - \mu_i
\end{IEEEeqnarray}

\noindent Recall that the UCB strategy for stochastic bandit can be expressed as follows:
\begin{IEEEeqnarray}{lCr}
    UCB_i(t-1, \gamma) & = & 
    \begin{cases}
    \infty , & N_{i, t-1} = 0\\
    \hat{x}_{i, t-1} + c\sqrt{\frac{\ln(1 / \gamma)}{N_{i, t-1}}}, & otherwise
    \end{cases}
\end{IEEEeqnarray}

\noindent Here $c$ is a constant, e.g., $c=\sqrt{2}$. Then, we can reach standard UCB by setting $\gamma = \frac{1}{t-1}$.
\begin{IEEEeqnarray}{lCr}
    UCB_i(t-1, \frac{1}{t-1}) & = & 
    \begin{cases}
    \infty , & N_{i, t-1} = 0\\
    \hat{x}_{i, t-1} + c\sqrt{\frac{\ln(t-1)}{N_{i, t-1}}}, & otherwise
    \end{cases}
\end{IEEEeqnarray}

\noindent For our UA-UCB, we can reach its exact format by using $\gamma = \frac{1}{{(t-1)}^{\beta_i}}$, and $\beta_i = c^2(1-\delta_i)^2$:
\begin{IEEEeqnarray}{lCr}
\label{eq:uaucb}
    UCB_i(t-1, \frac{1}{{(t-1)}^{\beta_i}}) & = & 
    \begin{cases}
    \infty , & N_{i, t-1} = 0\\
    \hat{x}_{i, t-1} + \sqrt{\frac{\beta_i \ln(t-1)}{N_{i, t-1}}}, & otherwise
    \end{cases}
\end{IEEEeqnarray}

\noindent Then the action at time $t$ is $A_t = \arg\max_i UCB_i(t-1,\gamma)$.  We simplify the second term as $c_{t-1, N_{i,t-1}}$, and show our UAUCB has a tighter regret bound with $\beta_i$.\\

\noindent Proof.
\begin{IEEEeqnarray}{lCl}
\label{eq:count}
N_{i, n} &=& \sum_{t=1}^n \mathbbm{1}\{ A_t = i \} \nonumber \\
        &=& \sum_{t=1}^k \mathbbm{1}\{ A_t = i \} +  \sum_{t=k+1}^n \mathbbm{1}\{ A_t = i \} \nonumber \\
        &=& 1+ \sum_{t=k+1}^n \bigg(\mathbbm{1}\{ A_t = i, N_{i, t-1} \geq l \} + \mathbbm{1}\{ A_t = i, N_{i, t-1} < l \}\bigg) \nonumber \\
        &=& 1+ \sum_{t=k+1}^n \mathbbm{1}\{ A_t = i, N_{i, t-1} \geq l \} + \sum_{t=k+1}^n \mathbbm{1}\{ A_t = i, N_{i, t-1} < l \} \nonumber \\
        &\leq & l + \sum_{t=k+1}^n \mathbbm{1}\{ A_t = i, N_{i, t-1} \geq l \} \nonumber \\
        & & (A_t = i \iff \forall j \in {1, 2, ..., k}: UCB_j \leq UCB_i \Rightarrow UCB_{\hat{i}^*} \leq UCB_i) \nonumber \\
        &\leq & l + \sum_{t=k+1}^n \mathbbm{1}\{UCB_{\hat{i}^*} \leq UCB_i, N_{i, t-1} \geq l \} \nonumber \\
        &=& l + \sum_{t=k+1}^n \mathbbm{1}\{\hat{x}_{\hat{i}^*, N_{\hat{i}^*,t-1}} + c_{t-1, N_{\hat{i}^*,t-1}} \leq \hat{x}_{i, N_{i,t-1}} + c_{t-1, N_{i,t-1}}, N_{i, t-1} \geq l \}
\end{IEEEeqnarray}
$l$ is a positive integer we determine later. And now, either one of the three following inequalities must hold:
\begin{IEEEeqnarray}{lCl}
& (I) &  \hat{x}_{\hat{i}^*, N_{\hat{i}^*,t-1}}  \leq \hat{\mu}_{\hat{i}^*} - c_{t-1, N_{\hat{i}^*,t-1}} \nonumber \\
& (II) & \hat{x}_{i, N_{i,t-1}} \geq \hat{\mu}_{i} + c_{t-1, N_{i,t-1}} \nonumber \\
& (III) & \hat{\mu}_{\hat{i}^*} < \hat{\mu}_i + 2c_{t-1, N_{i,t-1}} \nonumber
\end{IEEEeqnarray}
Proof by contradiction, assume all of them do not hold:
\begin{IEEEeqnarray}{lCl}
&  (I), (II) & \text{  } \hat{x}_{\hat{i}^*, N_{\hat{i}^*,t-1}}  > \hat{\mu}_{\hat{i}^*} - c_{t-1, N_{\hat{i}^*,t-1}} \text{ ,  } \hat{x}_{i, N_{i,t-1}} < \hat{\mu}_{i} + c_{t-1, N_{i,t-1}} \nonumber \\
&  \Rightarrow & \text{  } \hat{\mu}_{\hat{i}^*} < \hat{x}_{\hat{i}^*, N_{\hat{i}^*,t-1}}  +  c_{t-1, N_{\hat{i}^*,t-1}}  \text{ ,  } \hat{x}_{i, N_{i,t-1}} + c_{t-1, N_{i,t-1}} < \hat{\mu}_{i} + 2c_{t-1, N_{i,t-1}} \nonumber \\
& \text{From Eq.\ref{eq:count} we have: } & \hat{x}_{\hat{i}^*, N_{\hat{i}^*,t-1}}  + c_{t-1, N_{\hat{i}^*,t-1}} \leq \hat{x}_{i, N_{i,t-1}} + c_{t-1, N_{i,t-1}} \nonumber \\
& \Rightarrow & \hat{\mu}_{\hat{i}^*} < \hat{x}_{\hat{i}^*, N_{\hat{i}^*,t-1}}  + c_{t-1, N_{\hat{i}^*,t-1}} \leq \hat{x}_{i, N_{i,t-1}} + c_{t-1, N_{i,t-1}} < \hat{\mu}_{i} + 2c_{t-1, N_{i,t-1}} \nonumber \\
&\Rightarrow& \hat{\mu}_{\hat{i}^*} < \hat{\mu}_{i} + 2c_{t-1, N_{i,t-1}} \rightarrow \text{  (III) holds}
\end{IEEEeqnarray}
Thus, either (I), (II), (III) hold.
Then we calculate the probability of equations (I) and (II).
\begin{IEEEeqnarray}{lCl}
\label{eq:ineq 1}
p(I) &=& p( \hat{x}_{\hat{i}^*, N_{\hat{i}^*,t-1}}  \leq \hat{\mu}_{\hat{i}^*} - c_{t-1, N_{\hat{i}^*,t-1}} ) \nonumber \\
 &=& p(  -\hat{x}_{\hat{i}^*, N_{\hat{i}^*,t-1}}  - (-\hat{\mu}_{\hat{i}^*}) \geq  c_{t-1, N_{\hat{i}^*,t-1}} ) \nonumber \\
 &\leq& e^{\frac{-2 N_{{\hat{i}^*,t-1}}^2 c_{t-1, N_{\hat{i}^*,t-1}}^2}{N_{\hat{i}^*,t-1} (0-1)^2}} \nonumber \\
 &=& e^{-2N_{\hat{i}^*,t-1} c_{t-1, N_{\hat{i}^*,t-1}}^2} \nonumber \\
 &=& e^{-2N_{\hat{i}^*,t-1} \frac{\beta_{\hat{i}^*} \ln (t-1)}{N_{\hat{i}^*,t-1}}} \nonumber \\
 &=& (t-1)^{-2 \beta_{\hat{i}^*}}
\end{IEEEeqnarray}
\begin{IEEEeqnarray}{lCl}
\label{eq: ineq 2}
p(II) &=& p( \hat{x}_{i, N_{i,t-1}} \geq \hat{\mu}_{i} + c_{t-1, N_{i,t-1}}  ) \nonumber \\
    &=& p(\hat{x}_{i, N_{i,t-1}} - \hat{\mu}_i \geq c_{t-1, N_{i,t-1}}) \nonumber \\
    &\leq& e^{\frac{-2 N_{i, t-1}^2 c_{t-1, N_{i,t-1}}^ 2}{N_{i, t-1} (1-0)^2}} \nonumber \\
    &=& e^{-2 N_{i, t-1} c_{t-1, N_{i,t-1}}^2} \nonumber \\
    &=& e^{-2 N_{i, t-1} \frac{\beta_i \ln (t-1)}{N_{i, t-1}}} \nonumber \\
    &=& (t-1)^{-2 \beta_{i}}
\end{IEEEeqnarray}

\noindent To make equation (III) do not hold, let  $l =  \lceil \frac{4\beta_i \ln (t-1)}{\hat{\Delta}_i ^ 2} \rceil$,
\begin{IEEEeqnarray}{lCl}
\hat{\mu}_{\hat{i}^*} - \hat{\mu}_i - 2c_{t-1, N_{i,t-1}} &=& \hat{\mu}_{\hat{i}^*} - \hat{\mu}_i - 2 \sqrt{\frac{\beta_i \ln (t-1)}{N_{i, t-1}}} \nonumber \\
&\geq & \hat{\mu}_{\hat{i}^*} - \hat{\mu}_i - 2 \sqrt{\frac{\beta_i \ln (t-1) \hat{\Delta}_i^2}{4 \beta_i \ln (t-1)}}\nonumber \\
&=& \hat{\mu}_{\hat{i}^*} - \hat{\mu}_i - \hat{\Delta}_i = 0 \nonumber \\
p(III) &=& p(\hat{\mu}_{\hat{i}^*} - \hat{\mu}_i - 2c_{t-1, N_{i,t-1}} < 0) = 0
\end{IEEEeqnarray}

\noindent Then we calculate the expected value of $N_{i, n}$ in Eq.\ref{eq:count}:
\begin{IEEEeqnarray}{lCl}
\mathbb{E}[N_{i, n}] & \leq & l + \mathbb{E} [\sum_{t=k+1}^n \mathbbm{1}\{UCB_{\hat{i}^*} \leq UCB_i, N_{i, t-1} \geq l \}] \nonumber \\
& = & l + \sum_{t=k+1}^n P(UCB_{\hat{i}^*} \leq UCB_i, N_{i, t-1} \geq l ) \nonumber \\
& \leq & \frac{4\beta_i \ln (n-1)}{\hat{\Delta}_i ^ 2} + \sum_{t=k+1}^n [4(t-1)^{-2\beta_i} +4(t-1)^{-2\beta_{\hat{i}^*}}]
\end{IEEEeqnarray}

\noindent Then we calculate the regret:
\begin{IEEEeqnarray}{lCl}
    R_n &=& \sum_{i=1}^{k} \mathbb{E}[N_{i,n}]\Delta_i \nonumber \\
    &\leq& \sum_{i=1}^{k} \Bigl[ \frac{4\beta_i \ln (n-1)}{\hat{\Delta}_i ^ 2} + \sum_{t=k+1}^n [4(t-1)^{-2\beta_i} +4(t-1)^{-2\beta_{\hat{i}^*}}]\Bigr] \Delta_i
\end{IEEEeqnarray}

\noindent We know:
\begin{IEEEeqnarray}{lCl}
    \Delta_i &=& \mu_{i^*} - \mu_i \nonumber \\
            &=& \mu_{i^*} - \mu_i + \hat{\mu}_{\hat{i}^*} - \hat{\mu}_i + \hat{\mu}_i -  \hat{\mu}_{\hat{i}^*}\nonumber \\
            &=&  \hat{\mu}_{\hat{i}^*} - \hat{\mu}_i + \hat{\mu}_i - \mu_i +\mu_{i^*} - \hat{\mu}_{\hat{i}^*} \nonumber \\
            &=& \hat{\Delta}_i +  \hat{\mu}_i - \mu_i +\mu_{i^*} - \hat{\mu}_{\hat{i}^*} \nonumber \\
            &\leq& \hat{\Delta}_i + \delta_i + \mu_{i^*} - \hat{\mu}_{\hat{i}^*}
\end{IEEEeqnarray}

\noindent Thus:
\begin{IEEEeqnarray} {lCl}
\label{eq:regret bound new}
    R_n &\leq&  \sum_{i=1}^{k} \Bigl[ \frac{4\beta_i \ln (n-1)}{\hat{\Delta}_i ^ 2} + \sum_{t=k+1}^n [4(t-1)^{-2\beta_i} +4(t-1)^{-2\beta_{\hat{i}^*}}]\Bigr](\hat{\Delta}_i + \delta_i + \mu_{i^*} - \hat{\mu}_{\hat{i}^*}) \nonumber \\
\end{IEEEeqnarray}

\noindent For $\sum_{t=k+1}^n [4(t-1)^{-2\beta_i} +4(t-1)^{-2\beta_{\hat{i}^*}}]$, $\beta_i = c^2(1-\delta_i)^2$, there are several cases depending on $\delta_i \in [0,1]$:

\noindent 1) if $\delta_i = 0$, there is no difference between $p_i$ and $\hat{p}_i$ and UA-UCB becomes normal UCB.

\noindent 2) if $0 < \delta_i < 1 - \sqrt{\frac{1}{2c^2}}$, we have
\begin{IEEEeqnarray}{lCl}
& \ &\delta_i  < 1 - \sqrt{\frac{1}{2c^2}} \nonumber \\
& \Rightarrow & 1 - \delta_i > \sqrt{\frac{1}{2c^2}} \nonumber \\
& \Rightarrow & c^2 (1-\delta_i)^2 > \frac{1}{2} \nonumber \\
& \Rightarrow & \beta_i > \frac{1}{2} \nonumber \\
& \Rightarrow & - 2\beta_i  <  -1
\end{IEEEeqnarray}

\noindent The series $\sum_{t=k+1}^n [4(t-1)^{-2\beta_i} +4(t-1)^{-2\beta_{\hat{i}^*}}]$ converge and assume the sum is smaller than constant $c_1$, from Eq.\ref{eq:regret bound new}, we get:
\begin{IEEEeqnarray}{lCl}
    R_n &\leq& \sum_{i=1}^{k} \Bigl[\frac{4\beta_i \ln (n-1)}{\hat{\Delta}_i^2} + c_1 \Bigr](\hat{\Delta}_i + \delta_i + \mu_{i^*} - \hat{\mu}_{\hat{i}^*}) \nonumber \\
    &= & \sum_{i=1}^{k} \frac{4\beta_i (\hat{\Delta}_i + \delta_i + \mu_{i^*} - \hat{\mu}_{\hat{i}^*}) \ln(n-1)}{\hat{\Delta}_i^2} + constant.
\end{IEEEeqnarray}

\noindent 3) if $\delta_i = 1 - \sqrt{\frac{1}{2c^2}}$, we have $-2\beta_i=-1$. Then $\sum_{t=k+1}^n 4(t-1)^{-2\beta_i}=\sum_{t=k+1}^n 4(t-1)^{-1}$ and it is part of the harmonic series $\sum_{t=1}^{n}t^{-1}$. Since
\begin{IEEEeqnarray}{lCl}
\sum_{t=1}^{n}t^{-1} \sim \ln n +\gamma,
\end{IEEEeqnarray}
where $\gamma \approx 0.577$ is the Euler–Mascheroni constant.
In this case, $\sum_{t=k+1}^n [4(t-1)^{-2\beta_i} +4(t-1)^{-2\beta_{\hat{i}^*}}]$ has the same order as $\ln (n-1)$, so $\sum_{t=k+1}^n 4(t-1)^{-2\beta_i}$ and $\frac{4\beta_i \ln (n-1)}{\hat{\Delta}_i ^ 2}$ are of the same order.

\noindent Assume $\sum_{t=k+1}^n [4(t-1)^{-2\beta_i} +4(t-1)^{-2\beta_{\hat{i}^*}}] \sim c_2 \ln (n-1)$, $c_2$ is a constant, 
then from Eq.\ref{eq:regret bound new}, we get: 
\begin{IEEEeqnarray}{lCl}
    R_n &\leq& \sum_{i=1}^{k} \Bigl[ \frac{4\beta_i \ln (n-1)}{\hat{\Delta}_i^2} + c_2 \ln (n-1) \Bigr] (\hat{\Delta}_i + \delta_i + \mu_{i^*} - \hat{\mu}_{\hat{i}^*}) \nonumber \\
    &=& \sum_{i=1}^{k} (\frac{4\beta_i}{\hat{\Delta}_i^2}+c_2)(\hat{\Delta}_i + \delta_i + \mu_{i^*} - \hat{\mu}_{\hat{i}^*})\ln (n-1) .
\end{IEEEeqnarray}

\noindent Thus, when $0 \leq \delta_i \leq \text{min}(1 - \sqrt{\frac{1}{2c^2}},1)$, the regret is bounded by $\ln (n-1)$, which is sublinear. 

\noindent \paragraph{Comparison with standard UCB:} the difference in regret bound lies on the exploration weight, i.e. $\beta_i = c^2$ for UCB. If $\delta_i$ increases, the bound of UCB increases due the the term $(\hat{\Delta}_i + \delta_i + \mu_{i^*} - \hat{\mu}_{\hat{i}^*})$. In contrast, in UA-UCB  there is $\beta_i = c^2(1-\delta_i)^2$, which makes the term $\frac{4\beta_i \ln (n-1)}{\hat{\Delta}_i^2}$ decrease, leading to a tighter bound than UCB.

\section{Proof for Lemma 1}

%\begin{proof}\renewcommand{\qedsymbol}{}
We prove the above lemma via contradiction.
    \begin{IEEEeqnarray}{lCl} \nonumber
        (\RomanNumeralCaps{1}) & \exists v_j \in Ch(v), a \in \mathbb{N}: \Lim{N_I\to\infty} N(v_j) < a & \nonumber \\
        & \Rightarrow \exists w \in \mathbb{N}, v_j \text{ will not be visited after iteration w.}& \nonumber
        % \\
        % & \text{visited after iteration } w.  & \nonumber
    \end{IEEEeqnarray}
    We will contradict $(\RomanNumeralCaps{1})$ and show that $v_j$ will get selected after iteration $w$.    
    \begin{IEEEeqnarray}{ll} \nonumber
            \Lim{N_I\to\infty}N(v)=\infty, \Lim{N_I\to\infty} N(v_j) < a ,0<\alpha_j<1 &\nonumber\\
            \Rightarrow \Lim{N_I\to\infty} \sqrt{\frac{\ln N(v)}{N(v_j)}}\cdot(1-\alpha_j) = \infty&\nonumber\\
            \Rightarrow \Lim{N_I\to\infty} \frac{Q(v_j)}{N(v_j)}+c\sqrt{\frac{\ln{N(v)}}{N(v_i)}} \cdot  (1 - \alpha_i) = \infty &\nonumber
    \end{IEEEeqnarray}
    Limiting the scope to episodic tasks:
    \begin{IEEEeqnarray}{lr} \nonumber
        \exists b \in \mathbb{R}^{+}, \forall v \in Tree: 0 \leq Q(v)/N(v) \leq b &  
    \end{IEEEeqnarray}
    Assume $v_i$ is the most visited child in $Ch(v)$:
    \begin{IEEEeqnarray}{lr} \nonumber
        \Rightarrow N(v_i)\geq N(v)/|Ch(v)| \Rightarrow \Lim{N_I=\infty} N(v_i)\to\infty & 
    \end{IEEEeqnarray}
    Now we show that UA-UCT value of $v_i$ is bounded:
    \begin{IEEEeqnarray}{lCl} \nonumber
        \Lim{N_I\to\infty} UA-UCT(v_i) &=& \Lim{N_I\to\infty} \frac{Q(v_i)}{N(v_i)}+c\sqrt{\frac{\ln N(v)}{N(v_i)}}\cdot(1-\alpha_i) \nonumber\\ 
        &\leq& \Lim{N_I\to\infty} b + c\cdot \sqrt{\frac{\ln N(v)}{N(v)/|Ch(v)|}}\cdot(1-\alpha_i)\nonumber\\
        &=&b+c\cdot 0\cdot (1-\alpha_i)\nonumber\\
        &=&b\nonumber
    \end{IEEEeqnarray}
    We showed that $\Lim{N_i\to\infty}UA-UCT(v_i) = b$ and $\Lim{N_i\to\infty} UA-UCT(v_j)=\infty$, thus $v_j$ will be chosen over $v_i$ when $N_I\to\infty$, and since $\Lim{N_I=\infty} N(v_i)\to\infty$, we conclude $v_j$ will be visited after iteration $w$. {\fontsize{24}{24}\selectfont \textreferencemark}
    % $\left\{
    % \begin{array}{lr}
    %     \Lim{N_I\to\infty}lnN(v)/N(v)=0\\
    %     0<\alpha_i<1
    % \end{array}
    % \right\}
    % \Rightarrow
    % \Lim{N_I\to\infty} UA-UCT(v_i) = b. \\
    % \Rightarrow \Lim{N_I\to\infty} UA-UCT(v_i) < UA-UCT(v_j)$.\\ 
%\end{proof}